\documentclass[11pt]{article}

\usepackage[final]{acl}
\usepackage{algorithm}
\usepackage{algpseudocode}
\usepackage{times}
\usepackage{latexsym}
\usepackage{booktabs}
\usepackage{amsmath} 
\usepackage[table]{xcolor}

\usepackage{graphicx}
\usepackage{multirow}
\usepackage{amsmath,amssymb}
\usepackage{listings}
\usepackage{xcolor}
\usepackage[most]{tcolorbox}

\usepackage{makecell}
\usepackage{natbib}
\usepackage[T1]{fontenc}

\usepackage[utf8]{inputenc}

\usepackage{microtype}

\usepackage{inconsolata}

\usepackage{graphicx}
\definecolor{stagebg}{RGB}{232,242,252} 

\newcommand{\stageline}[1]{%
    \Statex \colorbox{stagebg}{%
        \parbox{\dimexpr\linewidth-2\fboxsep\relax}{%
            \strut\textbf{#1}\strut
        }%
    }%
}
\title{H-OPD: Confidence Aware Heterogeneous Multi-Teacher Multimodal On-policy Distillation}

\author{
\textbf{Qixiang Yin\textsuperscript{1,5}},
\textbf{Huanjin Yao\textsuperscript{2}},
\textbf{Cai Yuchen\textsuperscript{3}},
\textbf{Jianghao Chen\textsuperscript{5}},
\textbf{Ziyi Wang\textsuperscript{1}},\\
\textbf{Fei Su\textsuperscript{1,4}},
\textbf{Zhicheng Zhao\textsuperscript{1,4,*}} \\
\textsuperscript{1} Beijing University of Posts and Telecommunications\quad
\textsuperscript{2} HKUST \quad \textsuperscript{3} USTC \\
\textsuperscript{4} Key Laboratory of Interactive Technology and Experience System, \\ Ministry of Culture and Tourism \quad
\textsuperscript{5} Zhongguancun Academy, Beijing, China \\ \textsuperscript{*} Corresponding Author 
\quad \quad \texttt{\{buptyqx,zhaozc\}@bupt.edu.cn} \\
 \\
}

\begin{document}
\maketitle
\begin{abstract}
On-policy distillation (OPD) has recently emerged as an effective post-training paradigm by providing supervision on student-generated trajectories. 
However, existing OPD methods for multimodal reasoning usually rely on a static teacher routing, assigning each sample to a single teacher based on modality or task type. This ignores that visual grounding and abstract reasoning may dominate different decoding steps, making a single teacher insufficient for the full trajectory.
To this end, H-OPD is proposed as a confidence-aware heterogeneous multi-teacher OPD framework for multimodal reasoning. By verifying the complementarity of heterogeneous teachers in the same reasoning process, H-OPD replaces task or sample level teacher routing with token-level teacher arbitration along the shared student trajectory. H-OPD employs vision-to-language description transfer to enable text-only teachers to access key visual semantics, and uses a confidence-aware arbitration mechanism to dynamically combine vision-language teacher and text-only teachers at each token. 
Extensive evaluations over 11 widely-used reasoning benchmarks showcase the superior performance of our method. Code is available at \url{https://github.com/buptyqx/H-OPD}.
\end{abstract}

\section{Introduction}
\label{Sec 1: Introduction}
\begin{figure}[htbp]
  \centering
  \includegraphics[width=\linewidth]{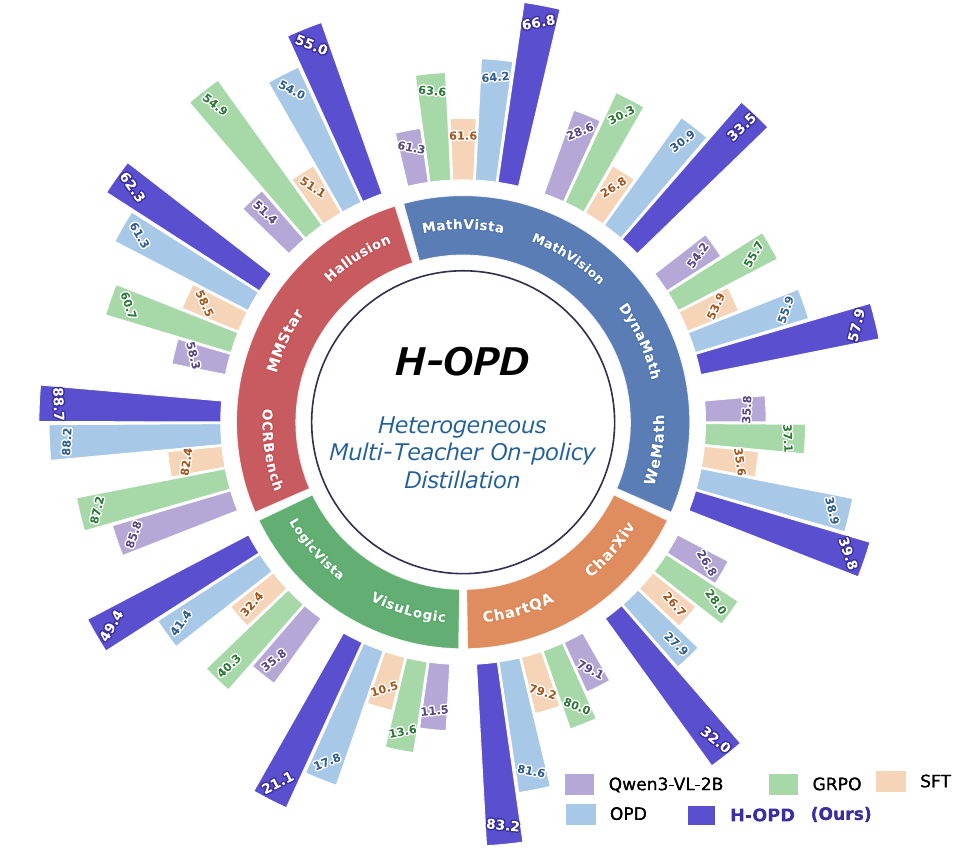}
  \caption{Performance comparison across 11 multimodal reasoning benchmarks. H-OPD consistently outperforms standard OPD and GRPO, leading to stronger overall performance.}
  \label{fig:first_page_fig_1}
\end{figure}

\begin{figure*}[ht]
  \centering
   \includegraphics[width=\linewidth]{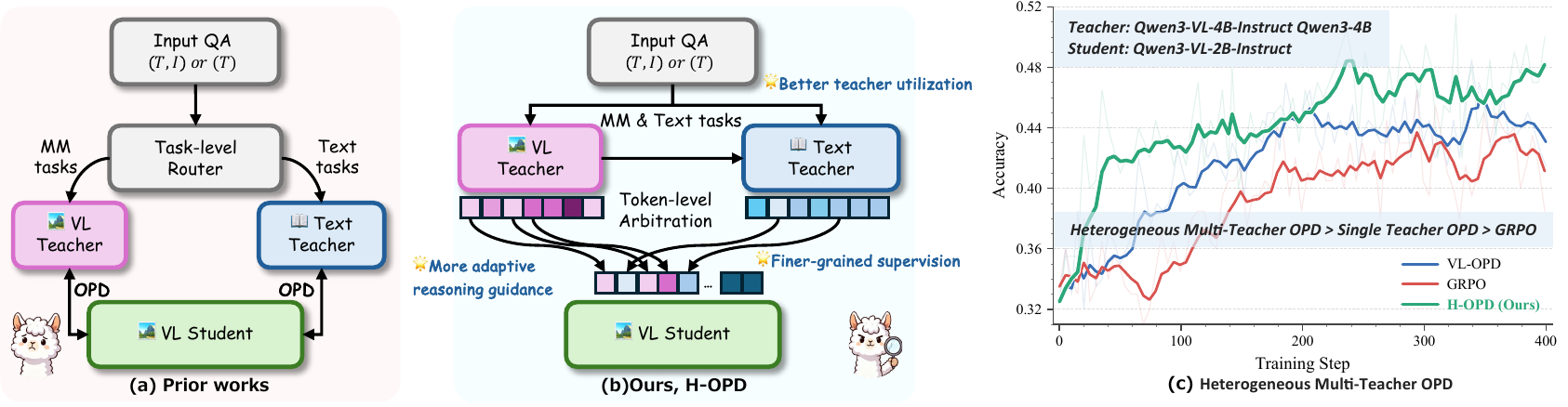}
   \caption{
Comparison between conventional task-level routing and our token-level arbitration OPD framework. (a) Conventional routing assigns each input to a single heterogeneous teacher based on task type. (b) Our method exposes the same input to both a VL teacher and a text teacher, and performs token-level arbitration along the shared student trajectory. (c) H-OPD consistently outperforms OPD and GRPO, highlighting the advantage of heterogeneous multi-teacher OPD.
}
   \label{figurepage2}
\end{figure*}

Effective post-training for multimodal reasoning remains an open challenge for MLLMs. 
Existing post-training frameworks typically follow a two-stage paradigm that combines supervised fine-tuning (SFT)~\citep{ouyang2022sft} with reinforcement learning (RL)~\citep{guo2025deepseekr1}. 
SFT relies on a fixed demonstration distribution and is therefore prone to exposure bias. 
RL can reduce distribution shift through online sampling, but it generally depends on sequence-level rewards and incurs much higher training costs. 
In contrast, on-policy distillation (OPD) provides efficient supervision on student-generated trajectories and has emerged as a promising paradigm for reasoning post-training. 
However, an effective framework for high-quality, fine-grained, and token-state-aware supervision is still lacking in multimodal reasoning.

As illustrated in Fig.~\ref{figurepage2}(a), existing approaches~\citep{lu2025onpolicydistillation_think_machine,yang2026self_distilled_RLVR} typically adopt relatively coarse-grained teacher routing strategies, they either train separate teacher-student systems for different modalities, or assign a single teacher to each input instance at the sample or task level, routing multimodal tasks to the vision-language(VL) teacher and text-only tasks to the text teacher. These methods commonly assume that the entire generation trajectory of a given input should be supervised by the same teacher. In fact, multimodal reasoning is not a static process governed by a single capability, but a dynamic autoregressive process with varying requirements for visual perception and abstract reasoning at different  decoding steps. Prior studies show that multimodal models can correctly perceive visual evidence yet still make reasoning errors, as perceptual information may be misinterpreted or overwritten during later reasoning process~\citep{liu2025more_thinking_less_seeing,wang2025papo,xu2026seeing_but_not_thinking}. This suggests that teacher reliability fluctuates along the response trajectory depending on token states. Accordingly, teacher allocation should be refined from the task or sample level down to the token level.

Furthermore, we argue that VL and text-only teachers are not redundant in multimodal reasoning; instead, they provide fine-grained complementary supervision throughout the reasoning trajectory. The former is more effective at perception-heavy steps, while the latter can be more reliable at reasoning-intensive stages. Therefore, instead of statically assigning a single teacher to each sample, it is more appropriate to expose the same input to multiple heterogeneous teachers and dynamically arbitrate their supervisory weights at each decoding token. Fig.~\ref{figurepage2}(b) illustrates this central idea: \textit{rather than performing task-level routing, our method performs token-level arbitration along the shared student trajectory.}

Motivated by these observations, we propose \textbf{H-OPD}, a \textbf{h}eterogeneous multi-teacher OPD framework for multimodal reasoning. 
The core idea of H-OPD is to feed the same input into both VL and text-only teachers, and perform token-level teacher arbitration rather than task/sample-level teacher selection.
Specifically, H-OPD first performs vision-to-language description transfer, converting key visual semantics of images into a textual proxy so that the text teacher lacking visual perception can also access task-relevant visual information in textual form. 
Building on this design, H-OPD further introduces a confidence-based token-level arbitration mechanism, which adaptively allocates supervisory weights according to the prediction entropy of each teacher at the current token, thereby enabling fine-grained collaborative distillation between visual grounding and abstract reasoning. 
As a result, heterogeneous teachers no longer compete as sample-level alternatives, but collaborate with each other at the token level. Empirically, as shown in Fig.~\ref{fig:first_page_fig_1} and Fig.~\ref{figurepage2}(c), H-OPD substantially outperforms conventional OPD and GRPO, achieving both higher training efficiency and superior test reasoning accuracy.
In summary, the main contributions of this work are as follows:

1. By exploring the transition from task/sample-level routing to token-level arbitration, we verify the complementarity of heterogeneous teachers in the same multimodal reasoning trajectory.

2. We propose H-OPD, to the best of our knowledge, the first heterogeneous multi-teacher on-policy distillation framework that enables token-level supervision by integrating complementary signals from VL and text-only teachers.

3. We conduct extensive experiments on 11 benchmarks, and the results  demonstrate the superiority of H-OPD across diverse multimodal reasoning tasks.

\section{Related Work}

\subsection{MLLM Reasoning}

Multimodal reasoning enhances slow-thinking capabilities~\citep{li2025system1system2}, enabling models to tackle complex reasoning tasks~\citep{lu2023mathvista}. Early work mainly relied on instruction tuning~\citep{liu2023llava} and reinforcement learning from human feedback (RLHF)~\citep{ouyang2022rlhf,yu2024rlhfv,yao2024mulberry}. However, scaling RLHF remains difficult due to its heavy dependence on high-quality human annotations~\citep{lee2023rlaif}. As a result, recent studies have explored more efficient RL-based strategies. Inspired by DeepSeek-R1~\citep{guo2025deepseekr1,cai2025predictability}, a line of work has introduced rule-based reward functions to provide reliable supervision without costly manual labeling~\citep{liu2025segzero,yao2025r1sharevl}. These methods design fine-grained rewards from multiple perspectives, including answer correctness~\citep{huang2025visionr1,yao2026mm_yao}, Chain-of-Thought completeness~\citep{zhang2025r1vl,cai2025predictability_rl}, and visual consistency~\citep{yang2025lookback,yin2026towards_yin}, improving the reasoning accuracy of MLLMs.

\subsection{On Policy Distillation}

Unlike traditional off-policy Knowledge Distillation (KD) that suffers from exposure bias resembling SFT~\citep{ouyang2022sft,lu2025onpolicydistillation_think_machine}, OPD~\citep{gu2024minillm,yang2026self_distilled_RLVR,hou2026uni_opd} aligns students on self-generated rollouts. By treating the teacher's token-level log-probabilities as dense rewards for RL algorithms like GRPO~\citep{yang2026exopd}, OPD efficiently bridges KD and RL~\citep{agarwal2024policy,cai2026learning2forsee}. Building on these insights, recent paradigms have expanded to self-distillation~\citep{zhao2026opsd,hubotter2026sdpo,ding2026hdpo} and multi-teacher settings~\citep{deepseekai2026deepseekv4,xiao2026mimo_v2_flash} to optimally fuse domain expert capabilities via unified framework~\citep{yang2026self_distilled_RLVR,yuan2026vision-opd}. Inspired by these breakthroughs, our approach achieves collaborative supervision from both VL and text-only teachers, improving multimodal reasoning in student models.

\section{Methodology}
The proposed \textbf{H-OPD} adaptively arbitrates between heterogeneous teachers along the student's trajectory, allowing complementary supervision from visual grounding and abstract reasoning. We first introduce the preliminaries in Sec.~\ref{sec:preliminaries}, then motivate token-level arbitration in Sec.~\ref{sec:motivation}, present the modality-bridging design in Sec.~\ref{sec:bridging}, and finally describe the confidence-aware arbitration objective in Sec.~\ref{sec:arbitration}.

\subsection{Preliminaries}
\label{sec:preliminaries}
\textbf{On-policy distillation.} OPD~\citep{gu2024minillm} addresses the exposure bias inherent in traditional off-policy KD by training the student model on its own generated trajectories rather than fixed teacher demonstrations. Given an input prompt $x$, the student policy auto-regressively generates a sequence $y \sim \pi_{\theta}(\cdot|x)$. At each decoding step $t$, the teacher model evaluates this self-generated state, providing a target token-level distribution $\pi_T = \pi_T(\cdot|x, y_{<t})$ as feedback for the student's parameterized distribution $\pi_{\theta} = \pi_{\theta}(\cdot|x, y_{<t})$. The student is then updated by minimizing the reverse KL divergence over its on-policy data: 
\begin{equation}
     \mathcal{L}(\theta) = \mathbb{E}_{x \sim \mathcal{D},\, y \sim \pi_{\theta}} \left[ \sum_{t=1}^{|y|} D_{\text{KL}}( \pi_{\theta}\,\|\,  \pi_T  ) \right] 
\end{equation}

By actively exploring the state space and receiving immediate corrective signals from the teacher on actual visited states, OPD ensures better train-inference consistency and robustly aligns the student's behavior.

\subsection{Motivation: From Task/Sample-level Routing to Token-level Arbitration}
\label{sec:motivation}
A common strategy for heterogeneous teacher distillation is to perform \emph{sample-level routing}: each input instance $x$ is assigned to a single teacher deemed most appropriate for that example.
For example, multimodal questions may be routed to a vision-language (VL) teacher, while text-only questions are routed to a text teacher.
Let $m^\star(x)\in\{1,\dots,M\}$ denote the selected teacher for input $x$.
Then the entire student trajectory for $x$ is supervised by the same teacher, $
\pi_{\mathrm{task}}^{(t)}(\cdot\mid x,y_{<t})
=
\pi_{m^\star(x)}(\cdot\mid x,y_{<t}).$ A slightly more general variant uses a fixed instance-level mixture over teachers:
\begin{equation}
\pi_{\mathrm{task}}^{(t)}(\cdot\mid x,y_{<t})
=
\sum_{m=1}^{M}\alpha_m(x)\,\pi_m(\cdot\mid x,y_{<t}),
\end{equation}
where, $\alpha(x)\in\Delta^{M-1}$ denotes an instance-specific mixing weight over the $M$ teachers. Hard routing is recovered as a special case when $\alpha(x)$ is one-hot form.

Despite their difference, both formulations assume that teacher authority is fixed for the entire input instance. However, OPD supervises the student on token states visited by its own on-policy rollout. Therefore, teacher reliability should depend on the current decoding state $s_t=(x,y_{<t})$, rather than only on the input instance $x$.

This motivates a new paradigm: \textit{instead of assigning one teacher to the whole sample, we expose the same input to multiple heterogeneous teachers and arbitrate their supervision at the token level.} Concretely, we define a state-dependent teacher mixture
\begin{equation}
\pi_{\mathrm{tok}}^{(t)}(\cdot\mid s_t)
=
\sum_{m=1}^{M}\lambda_m(s_t)\,\pi_m(\cdot\mid s_t).
\end{equation}
Here, $\lambda(s_t)\in\Delta^{M-1}$ denotes a state-dependent mixture weight over the $M$ teachers. Thus, multiple teachers may contribute to the same reasoning trajectory, while their relative authority can vary across decoding steps. Task-level routing is recovered as a special case when $\lambda_m(s_t)=\alpha_m(x)$ for all $t$; in particular, hard routing corresponds to one-hot $\alpha(x)$. Hence, sample-level routing and task-level fixed mixtures are both contained in the more general family of token-level arbitration.

Let $\pi^\star_t$ denote the ideal teacher distribution at state $s_t$.
Define $q_{\lambda,t}=\sum_{m=1}^{M}\lambda_m(s_t)\pi_{m,t}$ and $q_{\alpha,t}=\sum_{m=1}^{M}\alpha_m(x)\pi_{m,t}$, as the token-level and task-level teacher mixtures, respectively, where $\pi_{m,t}\equiv \pi_m(\,\cdot \mid s_t)$. Under any fixed distribution over visited token states, token-level arbitration admits an approximation error no worse than that of task-level routing:
\begin{equation}
\begin{aligned}
\inf_{\lambda}\,\mathbb{E}_{s_t}\!\left[
\mathrm{KL}\!\left(\pi^\star_t \,\|\, q_{\lambda,t}\right)
\right]
\le
\inf_{\alpha}\,\mathbb{E}_{s_t}\!\left[
\mathrm{KL}\!\left(\pi^\star_t \,\|\, q_{\alpha,t}\right)
\right].
\end{aligned}
\end{equation}
The inequality holds strictly when the most reliable teacher changes across token states within the same reasoning trajectory.
This is particularly plausible in multimodal reasoning, where different segments of the response rely on different capabilities, such as visual grounding, abstract reasoning, and linguistic fluency.

This raises a question: \textit{given that the VL teacher has access to the image, what additional benefit can a text-only teacher provide?}
Studies indicate that although multimodal models can accurately perceive visual evidence, they are prone to errors in reasoning, because perceptual information may be misinterpreted or overwritten during later inference~\citep{liu2025more_thinking_less_seeing,wang2025papo,xu2026seeing_but_not_thinking}.
Therefore, the VL teacher and the text-only teacher can complement each other both across instances and within the same reasoning process: the VL teacher offers stronger perceptual grounding, while the text-only teacher should achieve higher reliability at reasoning-heavy steps.

To verify this hypothesis, we conduct a pilot experiment on 100 questions sampled from MMFineReason~\citep{lin2026mmfinereason}, using \texttt{Qwen3-VL-4B-Instruct} as the VL teacher and \texttt{Qwen3-4B-Instruct-2507} as the text teacher.
We probe their token-level complementarity using two diagnostics.
The first is the top-$k$ Jaccard overlap,
$
\mathcal{J}^{(t)} =
\frac{|\mathcal{S}_{\mathrm{VL}}^{(t)} \cap \mathcal{S}_{\mathrm{Text}}^{(t)}|}
{|\mathcal{S}_{\mathrm{VL}}^{(t)} \cup \mathcal{S}_{\mathrm{Text}}^{(t)}|}
$,
which measures the overlap between their high-probability token sets at decoding step $t$.
The second is the entropy difference,
$
\Delta H^{(t)} = H_{\mathrm{VL}}^{(t)} - H_{\mathrm{Text}}^{(t)},
$
where positive values indicate higher uncertainty from the VL teacher and negative values indicate higher uncertainty from the text teacher.

\begin{figure}[t]
  \centering
  \includegraphics[width=\linewidth]{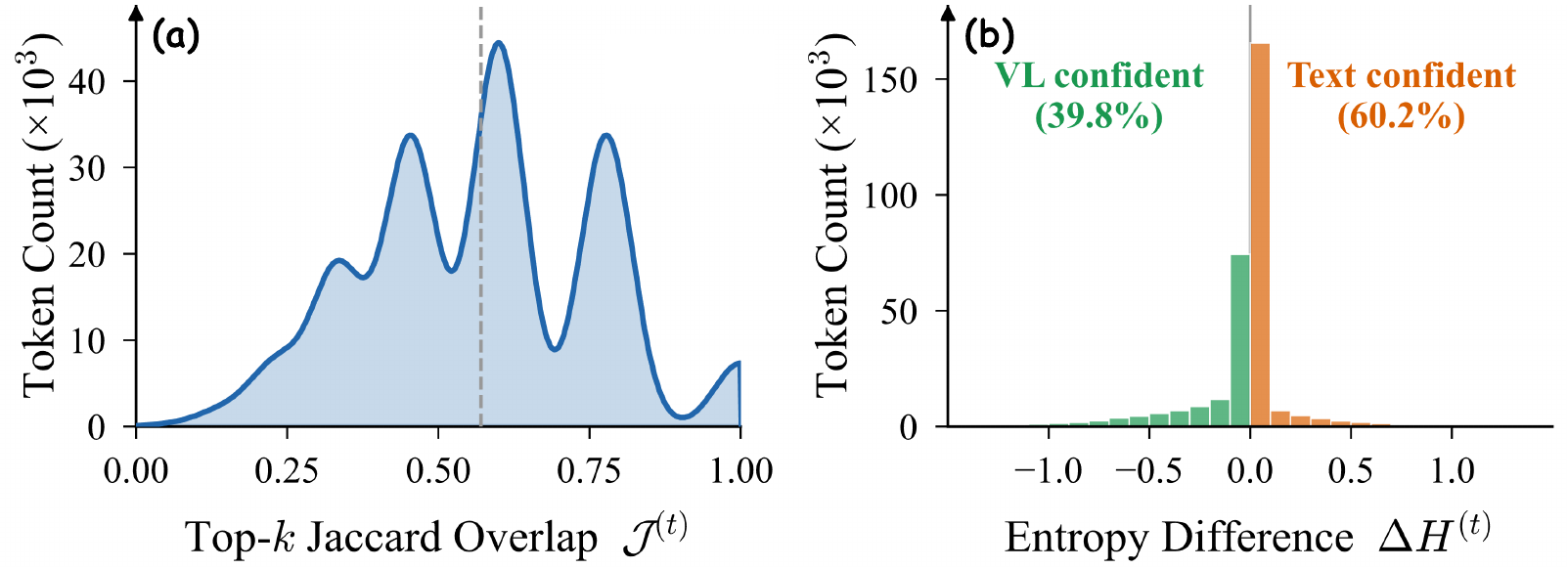}
  \caption{Pilot experiment on heterogeneous teacher distillation.
  (a) Distribution of top-$k$ Jaccard overlap between the VL teacher and the text teacher. 
  (b) Distribution of entropy difference, $\Delta H^{(t)} = H_{\mathrm{VL}}^{(t)} - H_{\mathrm{Text}}^{(t)}$. }
  \label{toy_exp}
\end{figure}

Fig.~\ref{toy_exp} supports this view.
The Jaccard overlap is far from concentrated near $1$, indicating that the two teachers tend to favor different candidate tokens at the same position.
Meanwhile, the entropy difference spans both positive and negative regions with substantial mass, showing that neither teacher is uniformly more confident throughout decoding.
Overall, these observations suggest that the two teachers provide complementary supervision within the same multimodal reasoning trajectory, thereby motivating token-level arbitration over sample-level routing.


\begin{figure*}[ht]
  \centering
   \includegraphics[width=\linewidth]{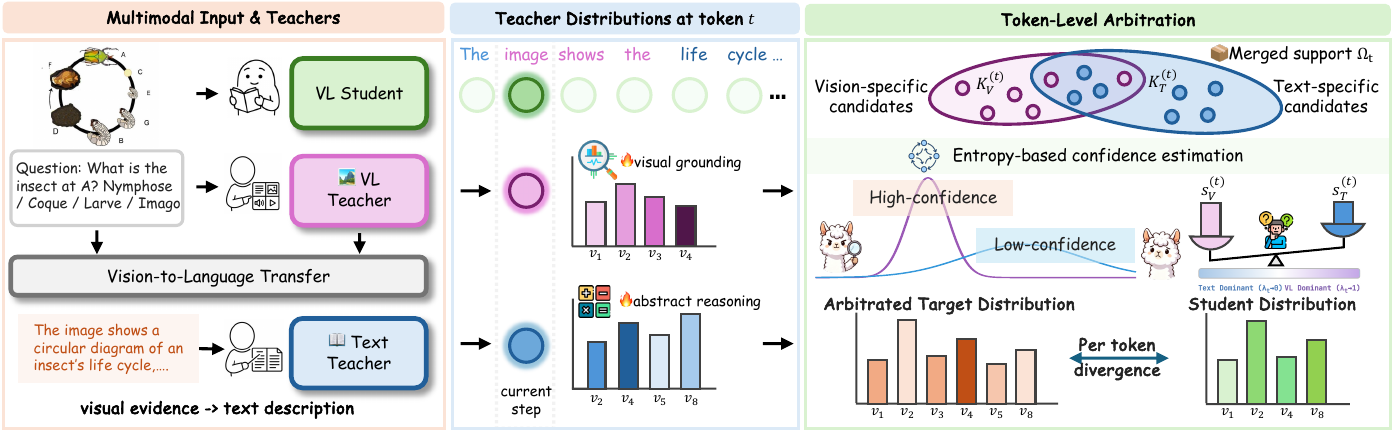}
\caption{Overview of H-OPD, which consists of a vision-to-language description transfer module for enabling the text-only teacher to access visual semantics, and a confidence-aware arbitration mechanism for dynamically integrating the vision-language teacher and the text-only teacher at the token level.}
   \label{fig:method_fig}
\end{figure*}

\subsection{Bridging the Modality Gap with Vision-to-Language Description Transfer}
\label{sec:bridging}
As shown in Fig.~\ref{fig:method_fig}, given a multimodal text-image input $(x, I)$, the student policy $\pi_\theta$ autoregressively generates an output sequence $y$, with the next-token distribution at step $t$ defined as $p_\theta^{(t)}(v) = \pi_\theta(v \mid x, I, y_{<t})$. Unlike standard online distillation relying on a single modality-aligned teacher, our framework employs two teachers with asymmetric modality access: a vision-language teacher $\tau_V$ conditioning on $(x, I)$, and a text-only teacher $\tau_T$ restricted to $x$. Since $\tau_T$ intrinsically lacks visual grounding, its direct application yields unfaithful supervision. Thus, the core challenge lies in effectively injecting visual context into $\tau_T$ while synergizing the complementary capabilities of both teachers during online training.

\textbf{Vision-to-Language Description Transfer.} Rather than forcing the text teacher to process visual features directly, we translate visual semantics into explicit textual proxies seamlessly consumable by $\tau_T$. 
Specifically, for a given sample $(x, I)$, we leverage the visionlanguage teacher to extract a task-relevant image description $d$. We then synthesize a modality-adapted prompt:
\begin{equation}
\tilde{x} = x^{\text{txt}} \oplus [\texttt{IMAGE DESCRIPTION}] \oplus d,
\end{equation}
where $x^{\text{txt}}$ denotes the original question text with the \texttt{<image>} placeholder removed, and $\oplus$ denotes string concatenation. This transformation disentangles the multimodal reasoning task by offloading visual \textit{perception} into the textual proxy $d$, thereby enabling the text-only teacher $\tau_T$ to focus exclusively on logical \textit{reasoning}. Because this perceptual stage is distilled into $d$ offline during data preprocessing, it introduces no additional serial overhead in the online rollout-and-distill loop.

\textbf{Heterogeneous Teacher and Student Policies.} Given the augmented prompt, the two teachers supervise the same student prefix through two parallel inference paths. 
At step $t$, the vision-language teacher produces $q_V^{(t)}(v) = \tau_V(v \mid x, I, y_{<t})$, 
while the text teacher produces
$q_T^{(t)}(v) = \tau_T(v \mid \tilde{x}, y_{<t})$.
To ensure tractability, we restrict the supervision to the top-$k$ predictions from each teacher. Let $K_V^{(t)}$ and $K_T^{(t)}$ denote the corresponding top-$k$ token sets, and let $\bar{q}_V^{(t)}$ and $\bar{q}_T^{(t)}$ denote the teacher distributions renormalized on their returned supports. The unified candidate vocabulary is defined as:
\begin{equation}
\Omega_t = K_V^{(t)} \cup K_T^{(t)}, \qquad |\Omega_t| \le 2k.
\end{equation}
When extending $\bar{q}_V^{(t)}$ and $\bar{q}_T^{(t)}$ to $\Omega_t$, tokens absent from a teacher's top-$k$ set are assigned zero probability. Crucially, conditioned on $y_{<t}$, the two teachers infer independently and asynchronously, bounding the per-step latency by the slower teacher alone.

\begin{table*}[t]
  \centering
  \renewcommand{\arraystretch}{1.08}
  \setlength{\tabcolsep}{4.6pt}
  \caption{
  Main results across multiple capability benchmarks.
  \texttt{Self} denotes the student model itself, Qwen3-VL-$x$B denotes \texttt{Qwen3-VL-$x$B-Instruct}.
  We report open-source checkpoints,and post-training results under different teacher configurations.$^*$ denotes evaluation on official weights using VLMEvalKit~\cite{duan2024vlmevalkit}.}
  \label{tab:qwen3vl_main}
  \resizebox{\textwidth}{!}{
  \begin{tabular}{lcccccccccccc}
    \toprule
    \multirow{2}{*}{\textbf{Setup (Teacher/Method)}}
    & \multicolumn{4}{c}{\textbf{Mathematics}}
    & \multicolumn{2}{c}{\textbf{Chart}}
    & \multicolumn{2}{c}{\textbf{Logic}}
    & \multicolumn{3}{c}{\textbf{General}}
    & \multirow{2}{*}{\textbf{Avg.}} \\
    \cmidrule(lr){2-5}
    \cmidrule(lr){6-7}
    \cmidrule(lr){8-9}
    \cmidrule(lr){10-12}
    & \thead{Math\\Vista}
    & \thead{Math\\Vision}
    & \thead{Dyna\\Math}
    & \thead{We\\Math}
    & \thead{CharXiv\\(RQ)}
    & \thead{Chart\\QA}
    & \thead{Visu\\Logic}
    & \thead{Logic\\Vista}
    & \thead{OCR\\Bench(/10)}
    & \thead{MMStar}
    & \thead{Hallusion\\Bench}
    & \\
    \midrule

    \multicolumn{13}{c}{\textit{\textbf{Baseline Models}}} \\
    \midrule
    Qwen3-VL-2B & 61.3 & 28.6$^*$ & 54.2   & 35.8$^*$ & 26.8 &  79.1 & 11.5  & 35.8 & 85.8 & 58.3 & 51.4 & 48.1 \\
    Qwen3-VL-4B & 73.7 & 44.7$^*$ & 65.3 & 54.3$^*$ & 39.7 & 84.6 & 19.0 & 53.2 & 88.1 & 69.8 & 57.6  & 59.1 \\
    Qwen3-VL-8B & 77.2 & 45.4$^*$ & 67.7  & 57.5$^*$ & 46.4 & 89.6  & 22.5 & 55.3 & 89.6 & 70.4 & 61.1 & 62.1 \\
    \midrule

    \multicolumn{13}{c}{\textit{\textbf{Student: Qwen3-VL-2B-Instruct}}} \\
    \midrule
    Self / GRPO ~\citep{guo2025deepseekr1} & 63.6 &  30.3& 55.7 & 37.1 & 28.0 & 80.0 & 13.6 & 40.3 & 87.2 & 60.7 & 54.9 & 50.1 \\
    \midrule
    Qwen3-VL-4B / SFT & 61.6 & 26.8 & 53.9 & 35.6 & 26.7 & 79.2 & 10.5 & 32.4 & 82.4 & 58.5 & 51.1 & 47.2 \\
    Qwen3-VL-4B / OPD~\citep{lu2025onpolicydistillation_think_machine} & 64.2 & 30.9 & 55.9 & 38.9 & 27.9 & 81.6 & 17.8 & 41.4 & 88.2 & 61.3 & 54.0 & 51.1 \\
    Qwen3-VL-4B / ExOPD~\citep{yang2026exopd} & 64.4 & 31.6 & 56.3 & 39.1 & 28.3 & 82.4 & 18.1 & 39.6 & 88.4 & 61.2 & 54.5 & 51.3 \\
    \rowcolor{blue!8}
    Qwen3-VL-4B + Qwen3-4B / H-OPD & \textbf{66.8} & \textbf{33.5} & \textbf{57.9} & \textbf{39.8} &\textbf{ 32.0 }& \textbf{83.2} & \textbf{21.1} &\textbf{49.4}  & \textbf{88.7 }& \textbf{62.3 }& \textbf{55.0} & \textbf{53.5}  \\

    \midrule
    Qwen3-VL-8B / SFT & 62.2 & 28.2 & 55.1 & 36.4 & 26.8 & 80.4 & 10.9 & 33.6 & 83.3 & 60.7 & 51.3 & 48.1 \\
    Qwen3-VL-8B / OPD~\citep{lu2025onpolicydistillation_think_machine} & 65.7 & 30.3 & 58.1 & 38.3 & 32.0 & 81.2 & 16.2 & 42.5 & 88.0 & 62.0 & 54.9 & 51.7 \\
    Qwen3-VL-8B / ExOPD~\citep{yang2026exopd} & 66.1 & 30.6 & 58.3 & 38.6 & 31.9 & 82.0 & 16.8 & 43.6 & 87.2 & 62.1 & 55.1 & 52.0 \\
    \rowcolor{blue!8}
    Qwen3-VL-8B + Qwen3-8B / H-OPD & \textbf{67.2} & \textbf{36.8} & \textbf{59.5} & \textbf{39.7} &\textbf{ 33.6} & \textbf{84.8} & \textbf{20.8 }& \textbf{50.3} & \textbf{89.5} &\textbf{ 62.7} & \textbf{55.4} & \textbf{54.6 } \\
    \midrule

    \multicolumn{13}{c}{\textit{\textbf{Student: Qwen3-VL-4B-Instruct}}} \\
    \midrule
    Self / GRPO ~\citep{guo2025deepseekr1}& 76.1 &  46.7& 67.9 & 56.2 & 40.9 & 88.0 & 21.0 & 54.1 & 89.1 & 70.0 & 60.2 & 60.9 \\
    \midrule
    Qwen3-VL-8B / SFT & 72.1 &  42.1& 65.5 & 54.3 & 39.6 & 84.8 & 18.1 & 51.5 & 87.7 & 68.0 & 55.4 & 58.1 \\
    Qwen3-VL-8B / OPD~\citep{lu2025onpolicydistillation_think_machine} & 76.6 & 46.1 & 68.3 & 55.7 & 40.5 & 88.8 & 22.4 & 53.5 & 88.2 & 68.8 & \textbf{60.9} & 60.9 \\
    Qwen3-VL-8B / ExOPD~\citep{yang2026exopd} & 77.2 & 46.4 & 68.9 &55.8  & 40.6 & 88.8 & 22.9 & 55.5 & 88.2 & 68.9 & \textbf{60.9} & 61.3 \\
    
    \rowcolor{blue!8}
    Qwen3-VL-8B + Qwen3-8B / H-OPD & \textbf{77.9} & \textbf{46.7} & \textbf{69.9 }& \textbf{56.3} & \textbf{41.0} & \textbf{88.9} & \textbf{23.1} &\textbf{ 55.9} & \textbf{89.3} & \textbf{70.0} & 59.3 & \textbf{61.7}  \\
    \bottomrule
  \end{tabular}}
\end{table*}

\subsection{Confidence-Aware Token-Level Teacher Arbitration}
\label{sec:arbitration}
The two teachers provide token-level supervision over the unified candidate set $\Omega_t$. Since different tokens rely on visual grounding and abstract reasoning to different extents, static ensembling is suboptimal, especially in the asymmetric modality setting. Confidence-Aware Token-Level Teacher Arbitration assigns token-wise supervisory emphasis according to predictive uncertainty. Teacher reliability is measured by predictive entropy, with higher entropy corresponding to greater uncertainty and lower entropy corresponding to higher confidence. For the renormalized teacher distributions over $\Omega_t$, the confidence scores are defined as:

\begin{equation}
s_V^{(t)} = -H\!\left(\bar{q}_V^{(t)}\right) \qquad
s_T^{(t)} = -H\!\left(\bar{q}_T^{(t)}\right)
\end{equation}
where $H(\cdot)$ denotes Shannon entropy. These confidence scores are then transformed into a token-level arbitration weight:
\begin{equation}
\lambda_t
=
\frac{\exp\!\left(s_V^{(t)} / \tau_f\right)}
{\exp\!\left(s_V^{(t)} / \tau_f\right)+\exp\!\left(s_T^{(t)} / \tau_f\right)}
\end{equation}
where $\tau_f > 0$ is a temperature parameter controlling the sharpness of arbitration. As $\tau_f \to \infty$, the weighting approaches uniform averaging; as $\tau_f \to 0$, it approaches hard selection of the more confident teacher. The resulting arbitrated target distribution is
\begin{equation}
q_A^{(t)}(v)
=
\lambda_t \bar{q}_V^{(t)}(v)
+
(1-\lambda_t)\bar{q}_T^{(t)}(v)
\end{equation}
where $v \in \Omega_t.$This construction enables supervision to adapt to the requirements of the current token, visually grounded tokens are influenced more strongly by $\tau_V$, whereas reasoning-dominant tokens place greater weight on $\tau_T$.

Given the arbitrated target distribution, the student policy is optimized by minimizing the reverse KL divergence over the restricted support:
\begin{equation}
\mathcal{L}(\theta)
=
\mathbb{E}_{y \sim \pi_\theta}
\left[
\sum_{t=1}^{|y|}
m_t \,
D_{\mathrm{KL}}^{\Omega_t}
\!\left(
p_\theta^{(t)} \,\|\,  q_A^{(t)} 
\right)
\right]
\end{equation}
where $m_t \in \{0,1\}$ denotes the mask over valid response positions, and $D_{\mathrm{KL}}^{\Omega_t}$ denotes the KL divergence evaluated only on $\Omega_t$.

Restricting distillation to the merged top-$k$ support improves both efficiency and stability. Since $|\Omega_t| \le 2k$, the per-step supervision cost is bounded by $\mathcal{O}(k)$, while the learning signal remains concentrated on the most relevant candidates.

\section{Experiments}

\subsection{Experiments Setup} 

\textbf{Training Dataset.} We train our models on MMFineReason-123K~\citep{lin2026mmfinereason}. Under the same experimental setting, we use a multimodal teacher model to generate image descriptions and employ GPT-4.1-mini~\citep{achiam2023gpt4} to assess the correctness of these descriptions. After filtering, we obtain 55K training samples, which are used consistently across all training settings.

\textbf{Models.} We conduct experiments on the Qwen3-VL~\citep{Qwen3-VL} model family. Specifically, we use Qwen3-VL-4B-Instruct and Qwen3-VL-2B-Instruct as student models. For heterogeneous teacher ensembles, we employ a combination of Qwen3-VL-4B-Instruct, Qwen3-4B-Instruct-2507, Qwen3-VL-8B-Instruct, and Qwen3-8B.

\textbf{Implementation Details.} We implement H-OPD based on the verl~\citep{sheng2025verl} framework, with vLLM~\citep{kwon2023vllm} used as the backend for both rollout generation and teacher inference. During training, the maximum prompt length and maximum response length are both set to 12,384 tokens. For teacher arbitration, we adopt confidence-aware fusion as the fusion strategy, with the fusion temperature $\tau_f$ set to 1.0. The student models are optimized using Adam with a learning rate of $1\times10^{-6}$ and a training batch size of 128. All experiments are conducted on a single compute node equipped with 8 NVIDIA B200 180GB GPUs.

\textbf{Evaluation Benchmarks.}
We assess H-OPD using a comprehensive benchmark suite spanning four major axes of multimodal capability. Specifically, for multimodal mathematical reasoning, we include MathVista~\citep{lu2023mathvista}, MathVision~\citep{wang2024mathvision}, DynaMath~\citep{zou2025dynamath}, and WeMath~\citep{qiao2025wemath}; for multimodal document understanding, we consider CharXiv~\citep{wang2024charxiv} and ChartQA~\citep{masry2022chartqa}; for multimodal logical reasoning, we evaluate on VisuLogic~\citep{xu2025visulogic} and LogicVista~\citep{xiao2024logicvista}; and for general multimodal visual question answering, we use OCRBench~\citep{liu2024ocrbench}, MMStar~\citep{chen2024mmstar}, and HallusionBench~\citep{guan2024hallusionbench}. Together, these benchmarks provide a broad and systematic assessment of H-OPD across a wide range of multimodal reasoning tasks.

\subsection{Main Results}

To provide a comprehensive evaluation of H-OPD, we conduct experiments on models of different scales, including 2B, 4B, and 8B variants. As shown in Table~\ref{tab:qwen3vl_main}, we organize the analysis into two parts. We first compare H-OPD against representative post-training baselines, including GRPO~\citep{guo2025deepseekr1}, OPD~\cite{lu2025onpolicydistillation_think_machine}, and ExOPD~\citep{yang2026exopd}. We then examine its scaling properties by analyzing the effects of increasing both student and teacher model sizes.

\textbf{Comparison with post-training strategies.} We first compare H-OPD against representative post-training strategies on the Qwen3-VL-2B-Instruct student model. As shown in Table~\ref{tab:qwen3vl_main}, H-OPD achieves a 5.5\% improvement in average score over the baseline instruct model. In addition, compared with OPD and ExOPD, which rely on distillation from single VL teacher, H-OPD improves the average score from 51.1 and 51.3 to 53.6 under the 4B teacher setting, yielding relative gains of 4.9\% and 4.5\%, respectively. The gains are particularly pronounced on challenging reasoning benchmarks such as MathVista, LogicVista, and CharXiv. This demonstrates that combining supervision from heterogeneous teacher allows H-OPD to transfer complementary multimodal perception and textual reasoning abilities more effectively, leading to superior multimodal reasoning performance over existing post-training baselines.

\textbf{Scaling up.} As model scale increases on both the teacher and student sides, H-OPD continues to show strong performance. For the Qwen3-VL-2B-Instruct student, replacing the 4B teacher pair with the 8B teacher pair improves the average score from 53.6 to 54.6, while maintaining a clear advantage over OPD and ExOPD under the same teacher configuration. This suggests that H-OPD can effectively benefit from stronger teachers. In addition, when applied to a larger Qwen3-VL-4B-Instruct student, H-OPD improves the average score from the baseline model’s 59.1 to 61.7, approaching the performance of its teacher Qwen3-VL-8B-Instruct.The improvements are particularly stable on multimodal mathematical reasoning and logical reasoning benchmarks, showing that H-OPD scales effectively on challenging cross-modal reasoning tasks. Overall, these results suggest that H-OPD benefits from both stronger teachers and larger students, highlighting its robustness as a general post-training framework for MLLM reasoning.

\subsection{Discussion}

\begin{figure}[t]
  \centering
  \includegraphics[width=\linewidth]{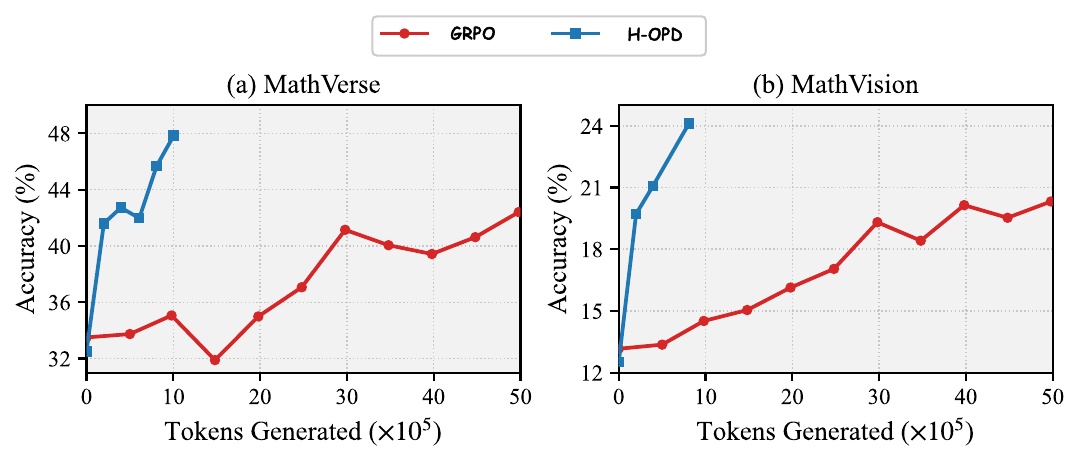}
  \caption{Comparison of training token efficiency between GRPO and H-OPD on (a) MathVerse and (b) MathVision. }
  \label{fig:token_eff}
\end{figure}

\textbf{Token Efficiency.} Figure~\ref{fig:token_eff} shows that H-OPD is substantially more token-efficient than GRPO. Across both MathVerse and MathVision, H-OPD attains higher accuracy within the first 100 training steps while using far fewer generated tokens, indicating faster convergence and more effective learning from each rollout.

\begin{figure*}[htbp]
  \centering
  \includegraphics[width=0.97\linewidth]{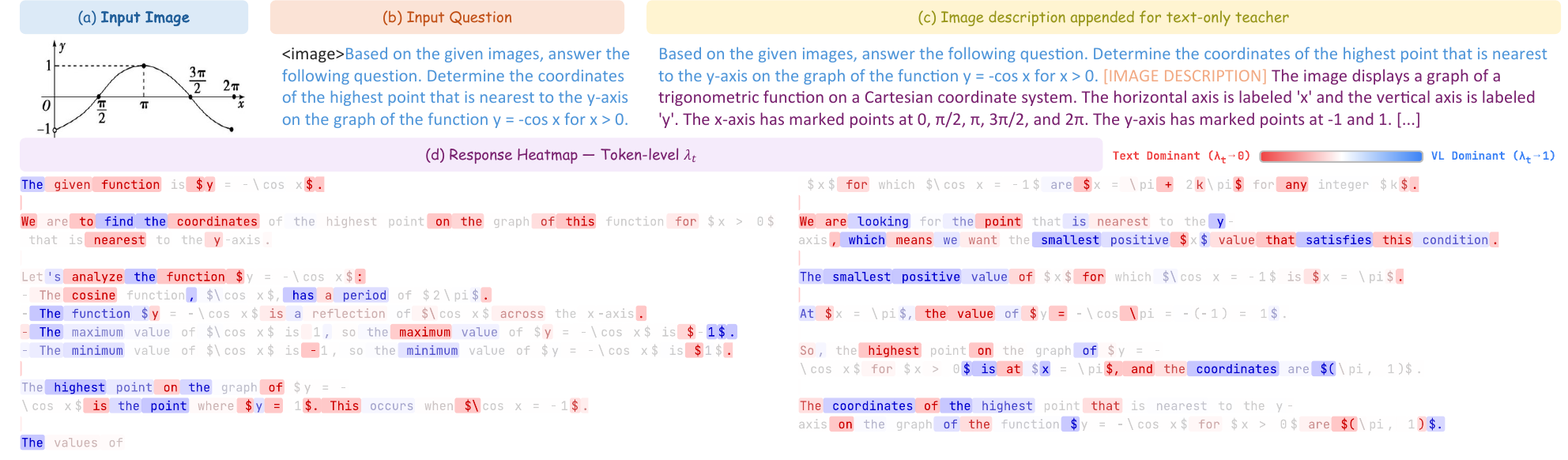}
  \caption{Visualization of the token-level teacher arbitration in H-OPD. (a) Input image. (b) Input question. (c) Textual description of the original image constructed for the text teacher. (d) Token-level heatmap of $\lambda_t$.}
  \label{fig:visfig}
\end{figure*}

\textbf{Stronger RL Teacher.} We further evaluate H-OPD with a stronger RL-trained teacher. Specifically, the teacher model is a version of Qwen3-VL-4B-Instruct trained with GRPO on the same training data. As shown in Table~\ref{tab:RL_H-OPD_RESULT}, compared with OPD and ExOPD, H-OPD improves the average score by 3.9 and 4.0 points, respectively. These results suggest that H-OPD can more effectively exploit the capabilities of a stronger RL-trained teacher, enabling better transfer of advanced reasoning skills to the student model.

\begin{table}[t]
  \centering
  \renewcommand{\arraystretch}{1.08}
  \setlength{\tabcolsep}{5.5pt}
\caption{Results with a stronger RL-trained teacher on selected benchmarks. Qwen3-VL-2B-Instruct is used as the student, and Qwen3-VL-4B-RL is a stronger teacher further trained with GRPO on the same data.}

  \label{tab:RL_H-OPD_RESULT}
  \resizebox{0.95\linewidth}{!}{
  \begin{tabular}{lcccc}
    \toprule
    \textbf{Method} & \textbf{MathVista} & \textbf{CharXiv} & \textbf{LogicVista} & \textbf{MMStar} \\
    \midrule
    Qwen3-VL-2B(S) & 61.3 & 26.8 & 35.8 & 58.3 \\
    Qwen3-VL-4B-RL(T) & 76.1 & 40.9 & 54.1 & 70.0 \\
    \midrule
    OPD & 68.8 & 32.4 & 42.5 & 63.3 \\
    ExOPD & 69.2 & 30.2 & 42.9 & 64.0 \\
    \rowcolor{blue!8}
    H-OPD(Ours) & \textbf{69.9} & \textbf{34.5} & \textbf{51.4} &\textbf{ 66.6} \\
    \bottomrule
  \end{tabular}}
\end{table}

\subsection{Ablation Study}
We conduct ablation studies with Qwen3-VL-2B-Instruct as the student and Qwen3-VL-4B-Instruct together with Qwen3-4B as the teachers. As shown in Table~\ref{tab:params_combined}, $\tau_f=0.5$ achieves the best performance on both MathVista and MathVision, indicating that a moderate fusion temperature provides the best balance between the two teacher signals. For top-$k$, setting $k=8$ offers a good trade-off between performance and efficiency, while a larger value brings only marginal gains at higher cost.

\begin{table}[t]
\centering
\caption{Ablation studies on key hyperparameters in H-OPD.}
\label{tab:params_combined}

\begin{minipage}[t]{0.48\columnwidth}
\centering
\small
\textbf{(a) The impact of $\tau_f$}
\vspace{2mm}

\resizebox{\textwidth}{!}{
\begin{tabular}{lcc}
\toprule
\textbf{$\tau_f$} & \textbf{MathVista} & \textbf{MathVision} \\
\midrule
0.1 & 66.1 & 31.0 \\
\rowcolor{blue!8}
0.5 & 66.8 & 33.5 \\
1.0 & 66.2 &  32.9 \\
1.5 & 66.2 & 32.3 \\

\bottomrule
\end{tabular}
}
\end{minipage}
\hfill
\begin{minipage}[t]{0.48\columnwidth}
\centering
\small
\textbf{(b) The impact of top-$k$}
\vspace{2mm}

\resizebox{\textwidth}{!}{
\begin{tabular}{lcc}
\toprule
\textbf{$k$} & \textbf{MathVista} & \textbf{Time/step} \\
\midrule
4  & 66.2 & 62 s \\
\rowcolor{blue!8}
8  & 66.8 & 92 s \\
32 & 67.1 & 718 s \\
\bottomrule
\end{tabular}
}
\end{minipage}

\end{table}

\subsection{Qualitative Evaluation}

To intuitively demonstrate the effectiveness of the token-level teacher arbitration coefficient $\lambda_t$ in H-OPD, we visualize its behavior with a token-level heatmap. As shown in Figure~\ref{fig:visfig}(c), we first present the textual description of the original image generated via Vision-to-Language Description Transfer, which is used to construct the rollout trajectory for the text teacher. Figure~\ref{fig:visfig}(d) further shows the token-wise values of $\lambda_t$, where color indicates whether each token is predominantly dominant by the text teacher or the vision-language teacher. From the visualization, we observe a clear functional specialization between the two teachers: the vision-language teacher mainly supervises tokens associated with visual perception, while the text teacher primarily guides tokens related to logical reasoning. This observation verifies that H-OPD delivers complementary rather than redundant supervision signals, allowing distinct teacher models to guide different aspects of the student’s capabilities.

\section{Conclusion}

We propose H-OPD, a confidence-aware heterogeneous multi-teacher on-policy distillation framework for multimodal reasoning. By shifting from task and sample-level teacher routing to token-level teacher arbitration, H-OPD enables a vision-language teacher and a text-only teacher to provide complementary supervision within the same reasoning trajectory. Through vision-to-language description transfer and confidence-aware arbitration, H-OPD effectively combines visual grounding and abstract reasoning signals during student generation. Experiments on 11 multimodal benchmarks show that H-OPD consistently outperforms strong post-training baselines and achieves higher token efficiency, retaining efficacy with advanced teachers and larger student models.

\section{Limitations}
In this section, we discuss the limitations of our method. First, the effectiveness of the text-only teacher depends on the quality of the transferred image description used to bridge the modality gap, when this description is incomplete, noisy, or weakly aligned with the task, the resulting supervision may be suboptimal. Second, our arbitration uses predictive entropy over truncated top-$k$ distributions as a proxy for teacher reliability, but low entropy does not always imply token-level correctness, especially when teachers are overconfident or poorly calibrated. While merged top-$k$ supervision improves efficiency, multi-teacher querying still introduces additional training overhead, and truncation may discard informative low-probability candidates. In addition, due to limited computational resources, we do not evaluate our method on larger-scale settings such as MoE models or 32B-level dense models. Although we discuss scalability in the main text, further validation at larger scales remains important future work.

\section{Acknowledgments}

This work was supported by the Key Research and Development Program of Xizang (XZ202601ZY0042), Beijing Advanced Innovation Center for Future Blockchain and Privacy Computing (GJJ-26) and the BUPT Innovation and Entrepreneurship Support Program (2025-YC-T033).

\bibliography{custom}
\newpage
\appendix
\newpage
\section*{Appendix}

\section{Benchmark}

\subsection{Multimodal Mathematical Reasoning}
\begin{itemize}
    \item \textbf{MathVista}~\citep{lu2023mathvista}: MathVista evaluates the comprehensive mathematical reasoning capabilities of models within complex visual contexts, such as geometric figures, function plots, scientific diagrams, and natural images. In this paper, we utilize the mathvista MINI subset, which contains 1,000 questions.
    \item \textbf{MathVision}~\citep{wang2024mathvision}: MathVision is a visual mathematical reasoning benchmark specifically designed for real-world mathematics competition-level problems. It covers 16 distinct mathematical disciplines categorized into 5 difficulty levels. In our evaluation, we use the mathvision MINI subset, comprising 304 questions.
    \item \textbf{DynaMath}~\citep{zou2025dynamath}: DynaMath evaluates the robustness of vision-language models in mathematical reasoning by dynamically modifying visual numerical values or function-graph features within the same problem. The benchmark includes 501 high-quality seed questions across multiple topics, each expanded into 10 variants, for a total of 5,010 questions.
    \item \textbf{WeMath}~\citep{qiao2025wemath}: WeMath aims to explore and evaluate the human-like problem-solving processes of large language models. WeMath deconstructs complex visual mathematical problems into step-wise sub-problems to deeply investigate the knowledge application and underlying logic during the model's reasoning process. In our experiments, a total of 1,740 questions from this benchmark are utilized.
\end{itemize}

\subsection{Document/Chart Understanding}

\begin{itemize}
    \item \textbf{CharXiv (RQ)}~\citep{wang2024charxiv}: CharXiv (RQ) is a benchmark designed to evaluate the understanding of authentic and complex charts sourced from scientific literature. This dataset comprises 2,323 highly challenging chart question-answering pairs, manually curated and verified by domain experts.
    \item \textbf{ChartQA}~\citep{masry2022chartqa}: ChartQA is specifically tailored for chart question answering, demanding models to integrate visual feature extraction, data reading, and logical reasoning to answer complex questions concerning real-world charts. For our evaluation, we utilize the chartqa test split, which consists of 1,250 questions.
\end{itemize}

\subsection{Multimodal Logical Reasoning}
\begin{itemize}
    \item \textbf{VisuLogic}~\citep{xu2025visulogic}: VisuLogic is a purely vision-centric logical reasoning benchmark. It is dedicated to evaluating the native visual logical reasoning capabilities in spatial, attribute, and quantitative pattern recognition. The dataset contains 1,000 human-verified visual logic multiple-choice questions.
    \item \textbf{LogicVista}~\cite{xiao2024logicvista}: LogicVista evaluates the comprehensive logical cognitive abilities of models within everyday visual contexts. The benchmark consists of 448 meticulously designed multiple-choice questions, each accompanied by detailed, human-annotated reasoning processes.
\end{itemize}

\subsection{General Multimodal Visual Question Answering}
\begin{itemize}
    \item \textbf{OCRBench}~\cite{liu2024ocrbench}: OCRBench comprehensively assesses the OCR and text reasoning capabilities of multimodal models. The benchmark comprises 1,000 manually filtered questions.
    \item \textbf{MMStar}~\citep{chen2024mmstar}: MMStar is an elite-level vision-dependent benchmark, primarily introduced to mitigate the data leakage issue prevalent in traditional benchmarks. It contains 1,500 highly challenging questions subjected to rigorous human review, ensuring that each problem strictly necessitates visual content for its resolution.
    \item \textbf{HallusionBench}~\citep{guan2024hallusionbench}: HallusionBench is specifically developed to diagnose and evaluate language hallucination and visual illusion in multimodal large models. The benchmark comprises 346 images and 1,129 meticulously designed adversarial question-answering pairs.
\end{itemize}

\section{H-OPD Details}
\label{appendix:hopd_details}

\begin{algorithm*}[t]
\caption{H-OPD Training with Confidence-Aware Token-Level Arbitration}
\label{alg:hopd}
\begin{algorithmic}[1]
\Require Training set $\mathcal{D}=\{(x,I)\}$; student $\pi_\theta$; 
VL teacher $\pi_V$; text teacher $\pi_T$; top-$k$ size $k$; 
temperature $\tau_f$

\stageline{Stage I: Offline vision-to-language transfer}
\ForAll{$(x,I)\in\mathcal{D}$}
    \State Generate a task-relevant description $d$ from image $I$ using $\pi_V$
    \State Construct $\tilde{x} \gets x^{\mathrm{txt}}
    \oplus [\texttt{IMAGE DESCRIPTION}] \oplus d$, and store $(x,I,\tilde{x})$
\EndFor

\stageline{Stage II: Online on-policy distillation}
\Repeat
    \State Sample a minibatch $\mathcal{B}\subset\mathcal{D}$
    \ForAll{$(x,I,\tilde{x})\in\mathcal{B}$}
        \State Roll out $y\sim\pi_\theta(\cdot\mid x,I)$ and initialize $\mathcal{L}\gets 0$
        \For{$t=1$ to $|y|$}
            \State Obtain student prediction 
            $\pi_\theta^{(t)}\gets\pi_\theta(\cdot\mid x,I,y_{<t})$
            \State Query teachers 
            $\pi_V^{(t)}\gets\pi_V(\cdot\mid x,I,y_{<t})$ and
            $\pi_T^{(t)}\gets\pi_T(\cdot\mid\tilde{x},y_{<t})$
            
            \State Extract teacher candidates 
            $K_j^{(t)}\gets\operatorname{TopK}(\pi_j^{(t)},k)$ for $j\in\{V,T\}$, 
            set $\Omega_t\gets K_V^{(t)}\cup K_T^{(t)}$
            
            \State Truncate and normalize each teacher distribution on its candidates:
            \[
            \bar{\pi}_j^{(t)}(v)\gets
            \frac{\pi_j^{(t)}(v)\mathbf{1}[v\in K_j^{(t)}]}
            {\sum_{u\in K_j^{(t)}}\pi_j^{(t)}(u)},
            \qquad j\in\{V,T\},\; v\in\Omega_t
            \]
            
            \State Measure teacher confidence by entropy:
            $\;s_j^{(t)}\gets-H(\bar{\pi}_j^{(t)})$, for $j\in\{V,T\}$
            
            \State Compute the confidence-aware arbitration weight:
            $\displaystyle
            \lambda_t\gets
            \frac{\exp(s_V^{(t)}/\tau_f)}
            {\exp(s_V^{(t)}/\tau_f)+\exp(s_T^{(t)}/\tau_f)}$
            
            \State Fuse teacher predictions:
            $\displaystyle
            \pi_A^{(t)}\gets
            \lambda_t\bar{\pi}_V^{(t)}
            +(1-\lambda_t)\bar{\pi}_T^{(t)}$
            
            \State Accumulate token-level distillation loss:
            $\displaystyle
            \mathcal{L}\gets\mathcal{L}
            +m_t D_{\mathrm{KL}}^{\Omega_t}
            \bigl(\pi_\theta^{(t)}\,\|\,\pi_A^{(t)}\bigr)$
        \EndFor
        \State Update $\theta$ using $\nabla_\theta\mathcal{L}$
    \EndFor
\Until{convergence}
\end{algorithmic}
\end{algorithm*}

This section provides supplementary implementation details omitted from the main method description for clarity, with a focus on practical aspects of data construction, teacher querying, restricted-support distillation, and optimization. As shown in Algorithm~\ref{alg:hopd}, we present the full training pipeline and further elaborate on several necessary design details.

\subsection{Vision-to-Language Description Construction}
\label{appendix:v2l_details}

For each training sample, we first construct a text-only surrogate input for the text teacher by generating an image description with the VL teacher. In practice, this description is prompted to capture the visual information that is most relevant to downstream reasoning, rather than producing a generic caption. We use a concise instruction that encourages the model to summarize objects, quantities, spatial relations, text in the image, and other visually grounded cues that may affect the final answer. Prompt~\ref{box:v2l_prompt} presents the full prompt used in the transfer process. The generated description is then concatenated with the original textual question to form the text teacher input.

To reduce noise propagation, we keep the description-generation prompt fixed across all training samples and use deterministic decoding. We additionally truncate overly long descriptions to a predefined maximum length. This design stabilizes the transferred visual context throughout training and avoids introducing unnecessary variance into the text-teacher branch. Furthermore, after constructing the training data, we employ GPT-4.1-mini as a filter to remove samples whose generated descriptions are incorrect, inaccurate, or incomplete.

\begin{tcolorbox}[
    title=\textbf{Vision-to-Language Description Transfer Prompt},
    breakable,
    label={box:v2l_prompt}
]
\lstset{
    language=Python,
    basicstyle=\small\fontfamily{zi4}\selectfont,
    breaklines=true,
    breakindent=0pt,   
    columns=fullflexible,
    commentstyle=\color{black},
    keywordstyle=\color{black},
    frame=none,
    extendedchars=true,
    showstringspaces=false,
    inputencoding=utf8
}
\begin{lstlisting}[numbers=none]
You are an expert vision-to-text translator. Please observe the image carefully and follow these rules to provide an objective and exhaustive transcription:

1. For general natural scenes: Describe the content in extreme detail (subjects, actions, environment, colors, spatial relationships).
2. For mathematical or geometric problems: Rigorously describe the geometric structures, lines, coordinate axes, ALL numerical values, alphabetical labels, and math symbols.
3. For tables or structured data: Transcribe directly into Markdown table or LaTeX code.
4. For text/OCR: Transcribe verbatim and accurately.

Original Question:
{clean_question}

CRITICAL: DO NOT output any greetings, conversation, or "Okay". Output the pure visual description directly below.

\end{lstlisting}
\end{tcolorbox}

\subsection{On-Policy Rollout}
\label{appendix:teacher_querying}

Given a rollout $y=(y_1,\dots,y_T)$ from the student, at each step $t$ we query the teachers using the same prefix $y_{<t}$ so that all distributions are conditioned on the identical student state. This ensures that teacher feedback is aligned with the trajectory actually visited by the student. For the VL teacher, the input consists of the original multimodal sample; for the text teacher, the input consists of the question concatenated with the transferred image description. We compute teacher distributions independently at every decoding step, and no hidden states are shared across the two teachers. In our implementation, teacher outputs are obtained with standard next-token logits, followed by softmax normalization and top-$k$ truncation.

Since the two teachers may assign probability mass to different candidate sets, H-OPD performs distillation on the merged support $\Omega_t = K_V^{(t)} \cup K_T^{(t)}$. Let $\pi_\theta^{\Omega_t}(\cdot \mid s_t)$ denote the student distribution restricted to $\Omega_t$ and renormalized over this set:
\begin{equation}
    \pi_\theta^{\Omega_t}(v \mid s_t)=\frac{\pi_\theta(v \mid s_t)}{\sum_{u\in\Omega_t}\pi_\theta(u \mid s_t)}
\end{equation}

The arbitrated teacher target $\pi_A^{(t)}$ is already defined on $\Omega_t$. We then compute the stepwise restricted-support KL divergence as
\begin{equation}
\begin{aligned}
& D_{\mathrm{KL}}^{\Omega_t}
\!\left(
\pi_\theta^{(t)} \,\middle\|\, \pi_A^{(t)}
\right)
= \\
&\sum_{v\in\Omega_t}
\pi_\theta^{\Omega_t}(v\mid s_t)
\log
\frac{
\pi_\theta^{\Omega_t}(v\mid s_t)
}{
\pi_A^{(t)}(v\mid s_t)
}.
\end{aligned}
\end{equation}
This formulation avoids forcing the student to match teacher probabilities over the entire vocabulary, while still preserving competition among the candidate tokens that either teacher considers important.

\subsection{Support Merging and Zero-Filling}
\label{appendix:support_merge}

After obtaining the top-$k$ token sets $K_V^{(t)}$ and $K_T^{(t)}$, we construct the union support $\Omega_t$. Each teacher distribution is represented on $\Omega_t$ by assigning zero probability to tokens outside its own top-$k$ set. Formally, for a teacher-specific truncated distribution $\bar{\pi}_i^{(t)}$ with support $K_i^{(t)}$, we use
\begin{equation}
\tilde{\pi}_i^{(t)}(v)=
\begin{cases}
\bar{\pi}_i^{(t)}(v), & v\in K_i^{(t)},\\
0, & v\in \Omega_t \setminus K_i^{(t)},
\end{cases}
\end{equation}

where $i\in\{V,T\}$.
The final arbitrated distribution is then computed on the shared support using these zero-filled vectors. In practice, this implementation is simple and avoids introducing additional approximation beyond the top-$k$ truncation already used in the method.

The token-level mask $m_t$ is used to exclude positions that are not suitable for distillation. In particular, we set $m_t=0$ for padded positions and for decoding steps after the end-of-sequence token has already been produced. For valid response tokens, we set $m_t=1$ and apply the distillation loss normally.

Depending on the data format, certain special tokens can also be excluded from distillation if they do not carry semantic content for reasoning, such as template separators or formatting markers introduced by the chat prompt. This masking rule prevents such tokens from dominating the loss and makes the optimization more focused on content-bearing decisions.

\subsection{Optimization and Reproducibility Details}
\label{appendix:optimization_details}

We train the student with standard minibatch optimization and compute the H-OPD loss over all valid response positions in each sampled rollout. Teacher models remain frozen throughout training, and only the student parameters are updated. For reproducibility, Table~\ref{tab:hyperparameters} summarizes the experimental settings for distilling Qwen3-VL-2B-Instruct using Qwen3-VL-4B-Instruct and Qwen3-4B-Instruct-2507 as teacher models.

From a systems perspective, H-OPD is more computationally demanding than single-teacher OPD because each student rollout requires querying two teachers at every decoding step. To improve efficiency, we cache the transferred image descriptions offline before training and reuse them throughout all epochs. This reduces repeated multimodal preprocessing and makes the text teacher branch substantially cheaper during optimization.

\begin{table}[t] 
\centering
\caption{Hyperparameter settings for the training and distillation processes.}
\label{tab:hyperparameters}
\renewcommand{\arraystretch}{1.2} 
\resizebox{\columnwidth}{!}{
\begin{tabular}{llc}
\toprule
\begin{tabular}[c]{@{}l@{}}\textbf{Parameter} \\ \textbf{Category}\end{tabular} & \textbf{Hyperparameter} & \textbf{Value} \\
\midrule
\multirow{3}{*}{\begin{tabular}[c]{@{}l@{}}\textbf{Model} \\ \textbf{Architecture}\end{tabular}} 
& Student Model & Qwen3-VL-2B-Instruct \\
& VL Teacher Model & Qwen3-VL-4B-Instruct \\
& Text Teacher Model & Qwen3-4B-Instruct-2507 \\
\midrule
\multirow{4}{*}{\begin{tabular}[c]{@{}l@{}}\textbf{Data} \\ \textbf{Configuration}\end{tabular}} 
& Max Prompt Length & 12,384 \\
& Max Response Length & 12,384 \\
& Max Number of Tokens & 24,769 \\
& Train Batch Size & 128 \\
\midrule
\multirow{6}{*}{\begin{tabular}[c]{@{}l@{}}\textbf{Distillation} \\ \textbf{Settings}\end{tabular}} 
& Distillation Loss Mode & \texttt{kl\_topk} \\
& Top-$K$ (for KL divergence) & 8 \\
& Loss Max Clamp & 10.0 \\
& Log Prob Min Clamp & -10.0 \\
& Fusion Mode & \texttt{token-level} \\
& Fusion Temperature ($\tau$) & 0.5 \\
\midrule
\multirow{8}{*}{\begin{tabular}[c]{@{}l@{}}\textbf{Training} \\ \textbf{\&} \\ \textbf{Optimization}\end{tabular}} 
& Training Framework & verl~\citep{sheng2025verl} \\
& Rollout Framework & vLLM~\citep{kwon2023vllm} \\
& Learning Rate (Actor) & 1e-6 \\
& PPO Mini Batch Size & 128 \\
& Student Micro Batch Size per GPU & 1 \\
& Total Epochs & 15 \\
& Sequence Parallel Size & 1 \\
\midrule
\multirow{8}{*}{\begin{tabular}[c]{@{}l@{}}\textbf{Hardware} \\ \textbf{\&} \\ \textbf{Execution}\end{tabular}} 
& GPU Type & NVIDIA B200 \\
& Total Number of GPUs & 8 \\
& GPUs for Student Model & 4 \\
& GPUs for VL Teacher Model & 2 \\
& GPUs for Text Teacher Model & 2 \\
& GPU Memory Utilization & 0.8 \\
& FSDP Parameter Offload & True \\
& FSDP Optimizer Offload & True \\
\bottomrule
\end{tabular}
} 
\end{table}


\section{Additional Analyses and Discussion}
\label{app:additional_analysis}

This appendix provides additional analyses of Heterogeneous On-Policy Distillation (H-OPD), focusing on compute-normalized efficiency, the contribution of token-level teacher arbitration, robustness to imperfect visual descriptions, teacher-scale sensitivity, and the interpretation of teacher specialization. Unless otherwise noted, the student is Qwen3-VL-2B-Instruct, and the heterogeneous teacher pair consists of a vision--language (VL) teacher and a text-only teacher.

\subsection{Compute-Normalized Efficiency}
\label{app:compute_efficiency}

H-OPD evaluates two teacher distributions on each student-generated trajectory and therefore introduces additional teacher-side forward computation relative to single-teacher OPD. Nevertheless, this overhead should be considered jointly with the rollout cost and convergence rate. In particular, GRPO samples multiple responses per input ($N=8$ in our experiments) to estimate group-relative advantages, whereas OPD-style methods obtain a dense token-level learning signal from each student trajectory. We therefore compare GRPO, OPD, and H-OPD using wall-clock time and GPU-hours, in addition to training steps and generated tokens.

\paragraph{Best performance within 400 steps.}
Table~\ref{tab:compute_best_400} reports the best MathVerse-200 validation accuracy reached during the first 400 optimization steps and the compute required to reach that checkpoint. H-OPD attains the highest accuracy, improving over GRPO and OPD by $5.0$ and $3.0$ percentage points, respectively.

\begin{table}[t]
\centering
\caption{Best MathVerse-200 accuracy within the first 400 training steps. GPU-hours are computed using eight GPUs.}
\label{tab:compute_best_400}
\small
\setlength{\tabcolsep}{3.8pt}
\begin{tabular}{lrrrr}
\toprule
Method & Best Acc. & Step & Time (h) & GPU-h \\
\midrule
GRPO  & 46.5 & 285 & 23.7 & 190.0 \\
OPD   & 48.5 & 339 & 18.1 & 145.2 \\
H-OPD & \textbf{51.5} & 311 & 21.7 & 173.6 \\
\bottomrule
\end{tabular}
\end{table}

\paragraph{Cost-matched comparison.}
To avoid favoring methods with different per-step costs, Table~\ref{tab:cost_matched} compares the best-so-far accuracy under identical wall-clock and GPU-hour budgets. H-OPD is consistently competitive and yields the best result at every reported budget. The advantage is especially pronounced early in training, indicating that heterogeneous token-level supervision improves the utility of each rollout.

\begin{table*}[t]
\centering
\caption{Cost-matched MathVerse-200 results. Each entry is the best validation accuracy attained within the corresponding wall-clock and GPU-hour budget.}
\label{tab:cost_matched}
\small
\setlength{\tabcolsep}{6pt}
\begin{tabular}{rrrrrrr}
\toprule
\multicolumn{2}{c}{Compute budget} & \multicolumn{3}{c}{Best accuracy} & \multicolumn{2}{c}{H-OPD improvement} \\
\cmidrule(lr){1-2}\cmidrule(lr){3-5}\cmidrule(lr){6-7}
Time (h) & GPU-h & GRPO & OPD & H-OPD & vs. GRPO & vs. OPD \\
\midrule
 5 &  40 & 38.0 & 42.0 & \textbf{46.5} & +8.5 & +4.5 \\
10 &  80 & 38.5 & 45.0 & \textbf{47.0} & +8.5 & +2.0 \\
15 & 120 & 44.0 & 47.0 & \textbf{47.0} & +3.0 & +0.0 \\
20 & 160 & 44.0 & 48.5 & \textbf{49.5} & +5.5 & +1.0 \\
22 & 176 & 44.5 & 48.5 & \textbf{51.5} & +7.0 & +3.0 \\
\bottomrule
\end{tabular}
\end{table*}

\paragraph{Time to target accuracy.}
We additionally measure the compute required to reach fixed accuracy thresholds (Table~\ref{tab:time_to_accuracy}). H-OPD reaches all shared thresholds with less wall-clock time and fewer GPU-hours than both baselines. GRPO does not reach $48.0\%$, while only H-OPD reaches $50.0\%$ within the observed training window.

\begin{table}[t]
\centering
\caption{Wall-clock hours and GPU-hours required to reach a target MathVerse-200 accuracy. A dash denotes that the target was not reached.}
\label{tab:time_to_accuracy}
\small
\setlength{\tabcolsep}{3.2pt}
\resizebox{\columnwidth}{!}{%
\begin{tabular}{rccc}
\toprule
Target & GRPO & OPD & H-OPD \\
\midrule
40.0 & 10.83 / 86.67 & 3.58 / 28.60 & \textbf{1.40 / 11.20} \\
45.0 & 23.75 / 190.00 & 9.62 / 77.00 & \textbf{2.45 / 19.60} \\
48.0 & -- & 18.15 / 145.20 & \textbf{15.05 / 120.40} \\
50.0 & -- & -- & \textbf{21.70 / 173.60} \\
\bottomrule
\end{tabular}}
\end{table}

These results complement the token-efficiency analysis in the main paper. They do not imply that the second teacher is computationally free; rather, they show that its additional forward cost is offset by faster improvement under matched end-to-end compute budgets.

\subsection{Mechanism Analysis: Arbitration and Routing}
\label{app:arbitration_ablation}

We isolate the source of H-OPD improvements by comparing token-level confidence-aware arbitration with three alternatives: (i) uniform averaging of both teacher distributions at every token, (ii) sample-level routing that assigns one teacher to an entire response, and (iii) supervision from the text teacher alone after conditioning it on a VL-generated image description. All variants use the same Qwen3-VL-2B-Instruct student and the same Qwen3-VL-4B-Instruct and Qwen3-4B-Instruct-2507 teacher pair.

\begin{table}[t]
\centering
\caption{Controlled arbitration ablations on LogicVista.}
\label{tab:arbitration_ablation}
\small
\setlength{\tabcolsep}{4pt}
\begin{tabular}{lr}
\toprule
Method & Accuracy \\
\midrule
Student baseline & 35.8 \\
Uniform two-teacher averaging & 31.3 \\
Sample-level teacher routing & 42.5 \\
Text teacher only + description & 26.6 \\
H-OPD (token-level arbitration) & \textbf{49.4} \\
\bottomrule
\end{tabular}
\end{table}

Uniform averaging underperforms the student baseline, demonstrating that the benefit does not follow merely from exposing the student to two teachers. Heterogeneous teachers can assign high probability to incompatible continuations, and an unconditional average may blur both signals. Sample-level routing performs substantially better, confirming that selecting among heterogeneous teachers is useful, but remains $6.9$ points below token-level arbitration. This gap supports the central hypothesis that teacher reliability varies within a single multimodal trajectory as decoding alternates between perceptual grounding and symbolic reasoning.

The text-teacher-only variant performs poorly because textualized image evidence cannot replace direct supervision from a VL teacher. The vision-to-language transfer module should therefore be interpreted as an interface that enables a text teacher to participate in multimodal distillation, rather than as a substitute for visual supervision.

\section{The Use of Large Language Models (LLMs)}

In this paper, large language model (LLM)~\citep{team2023gemini} is only utilized to assist with the refinement of the English language. No LLMs are employed to generate new innovative ideas, and the research process is conducted by the researchers (humans). 

\section{Case studies}

\begin{figure*}[ht]
  \centering
   \includegraphics[width=\linewidth]{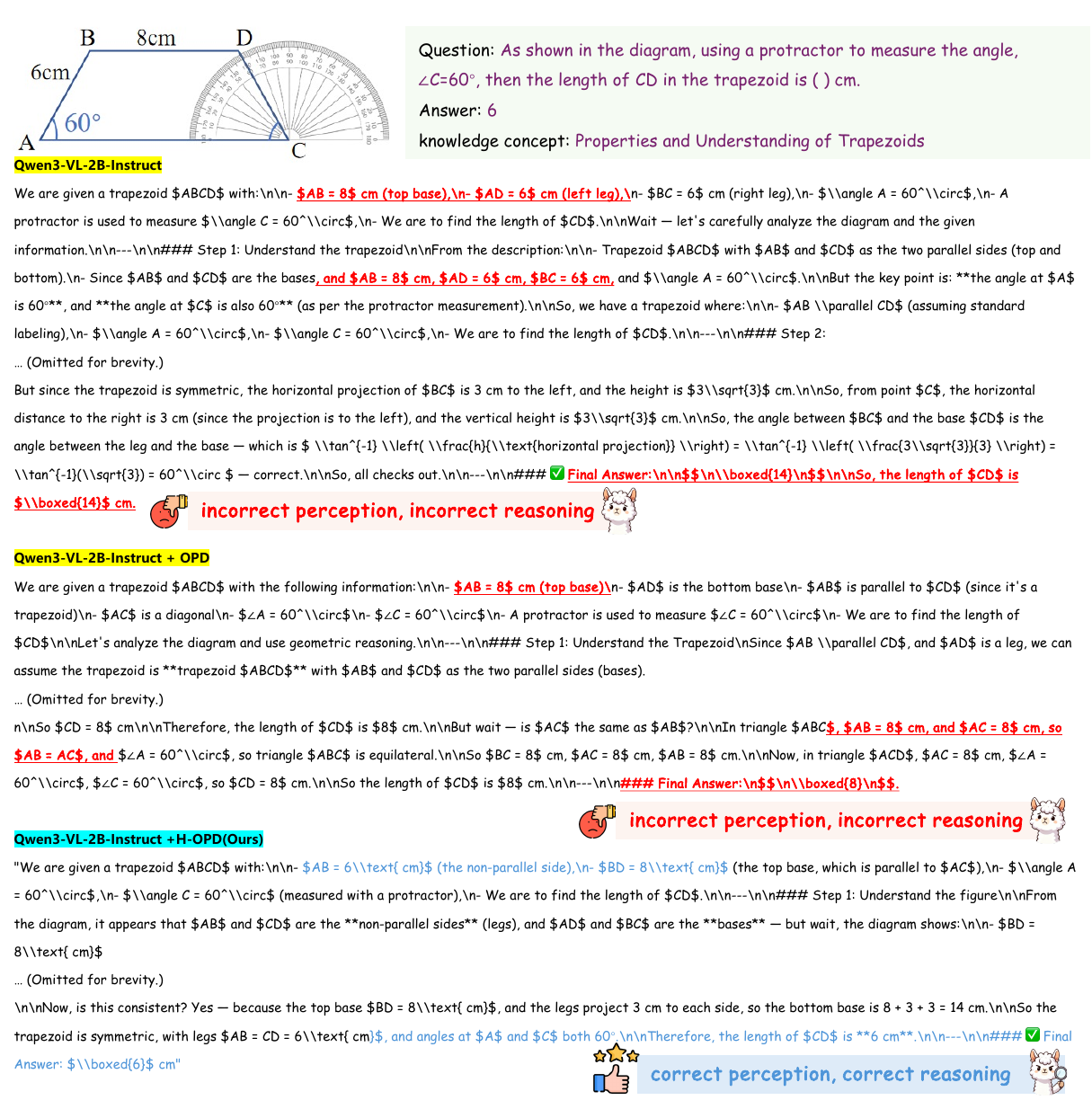}
   \caption{
Case studies.
}
   \label{case_study}
\end{figure*}

Fig.~\ref{case_study} presents a representative case study comparing the original model, the OPD baseline, and H-OPD on a geometry reasoning example. The original model and OPD both fail on this instance, but their errors stem from the same source: they first misinterpret the visual structure of the trapezoid and then build subsequent reasoning on top of that incorrect perception. Specifically, both models confuse the roles of the parallel sides and non-parallel sides, leading to an incorrect geometric decomposition and ultimately an erroneous final answer. This example shows that once the visual grounding is wrong, downstream reasoning can remain internally fluent yet still be fundamentally invalid.

In contrast, H-OPD correctly identifies the geometric configuration from the diagram and then follows a consistent reasoning process to derive the correct answer. This behavior highlights the benefit of heterogeneous token-level arbitration: the VL teacher contributes more reliable visually grounded signals when precise perception is required, while the text-side reasoning remains coherent after the correct structure has been established. The example therefore illustrates that the gain of H-OPD is not merely better final-answer accuracy, but a more faithful coupling between perception and reasoning. In multimodal reasoning tasks, such improvement is particularly important because reasoning quality is heavily constrained by the correctness of the underlying visual interpretation.

\end{document}